\documentclass[10pt,twocolumn,letterpaper]{article}

\usepackage{3dv}
\usepackage{times}
\usepackage{epsfig}
\usepackage{graphicx}
\usepackage{amsmath}
\usepackage{amssymb}
\usepackage{color}

\usepackage{color, colortbl}
\usepackage[pagebackref=true,breaklinks=true,letterpaper=true,colorlinks,bookmarks=false]{hyperref}

\threedvfinalcopy 

\def\threedvPaperID{95} 

\newcommand\blfootnote[1]{%
    \begingroup
    \renewcommand\thefootnote{}\footnote{#1}%
    \addtocounter{footnote}{-1}%
    \endgroup
}

\ifthreedvfinal\fi

\usepackage{fancyhdr}
\usepackage[headsep=1em,headheight=1.5em]{geometry} 
\definecolor{mygray}{gray}{0.6}
\begin{document}
\thispagestyle{plain}
\title{Deeper Depth Prediction with Fully Convolutional Residual Networks}

\author{
    Iro Laina$^{*1}$\\
    {\tt\small iro.laina@tum.de}
\and
Christian Rupprecht$^{*1,2}$\\
{\tt\small christian.rupprecht@in.tum.de}
\and
Vasileios Belagiannis$^{3}$\\
{\tt\small vb@robots.ox.ac.uk}
\and
Federico Tombari$^{1,4}$\\
{\tt\small tombari@in.tum.de}
\and
Nassir Navab$^{1,2}$\\
{\tt\small navab@in.tum.de}
}

\maketitle

\blfootnote{$^{*}$ equal contribution}
\blfootnote{$^{1}$ Technische Universit\"at M\"unchen, Munich, Germany}
\blfootnote{$^{2}$ Johns Hopkins University, Baltimore MD, USA}
\blfootnote{$^{3}$ University of Oxford, Oxford, United Kingdom}
\blfootnote{$^{4}$ University of Bologna, Bologna, Italy}

\begin{abstract}
This paper addresses the problem of estimating the depth map of a scene given a single RGB image. We propose a fully convolutional architecture, encompassing residual learning, to model the ambiguous mapping between monocular images and depth maps. In order to improve the output resolution, we present a novel way to efficiently learn feature map up-sampling within the network. For optimization, we introduce the reverse Huber loss that is particularly suited for the task at hand and driven by the value distributions commonly present in depth maps. Our model is composed of a single architecture that is trained end-to-end and does not rely on post-processing techniques, such as CRFs or other additional refinement steps. As a result, it runs in real-time on images or videos. In the evaluation, we show that the proposed model contains fewer parameters and requires fewer training data than the current state of the art, while outperforming all approaches on depth estimation. Code and models are publicly available$^{5}$.
\blfootnote{$^{5}$ \url{https://github.com/iro-cp/FCRN-DepthPrediction}} 
\end{abstract}

\section{Introduction}

Depth estimation from a single view is a discipline as old as computer vision and encompasses several techniques that have been developed throughout the years. One of the most successful among these techniques is Structure-from-Motion (SfM)~\cite{szeliski11}; it leverages camera motion to estimate camera poses through different temporal intervals and, in turn, estimate depth via triangulation from pairs of consecutive views. Alternatively to motion, other working assumptions can be used to estimate depth, such as variations in illumination~\cite{zhang1999shape} or focus~\cite{suwajanakorn2015}. 

In absence of such environmental assumptions, depth estimation from a single image of a generic scene is an ill-posed problem, due to the inherent ambiguity of mapping an intensity or color measurement into a depth value. While this also is a human brain limitation, depth perception can nevertheless emerge from monocular vision. Hence, it is not only a challenging task to develop a computer vision system capable of estimating depth maps by exploiting monocular cues, but also a necessary one in scenarios where direct depth sensing is not available or not possible. Moreover, the availability of reasonably accurate depth information is well-known to improve many computer vision tasks with respect to the RGB-only counterpart, for example in reconstruction~\cite{Silberman:ECCV12}, recognition~\cite{ren2012rgb}, semantic segmentation~\cite{Eigen15} or human pose estimation~\cite{taylor2012vitruvian}. 

For this reason, several works tackle the problem of monocular depth estimation. One of the first approaches assumed superpixels as planar and inferred depth through plane coefficients via Markov Random Fields (MRFs)~\cite{Saxena09}. Superpixels have also been considered in~\cite{Li15,Liu14,wang2015towards}, where Conditional Random Fields (CRFs) are deployed for the regularization of depth maps. Data-driven approaches, such as~\cite{Karsch12,Konrad12}, have proposed to carry out image matching based on hand-crafted features to retrieve the most similar candidates of the training set to a given query image. The corresponding depth candidates are then warped and merged in order to produce the final outcome.

Recently, Convolutional Neural Networks (CNNs) have been employed to learn an implicit relation between color pixels and depth~\cite{Eigen15,Eigen14,Li15,Liu15,wang2015towards}. CNN approaches have often been combined with CRF-based regularization, either as a post-processing step~\cite{Li15,wang2015towards} or via structured deep learning~\cite{Liu15}, as well as with random forests~\cite{roy16monocular}. These methods encompass a higher complexity due to either the high number of parameters involved in a deep network~\cite{Eigen15,Eigen14,Liu15} or the joint use of a CNN and a CRF~\cite{Li15,wang2015towards}. Nevertheless, deep learning boosted the accuracy on standard benchmark datasets considerably, ranking these methods first in the state of the art. 

In this work, we propose to learn the mapping between a single RGB image and its corresponding depth map using a CNN. The contribution of our work is as follows. 
First, we introduce a \emph{fully convolutional} architecture to depth prediction, endowed with novel up-sampling blocks, that allows for dense output maps of higher resolution and at the same time requires fewer parameters and trains on one order of magnitude fewer data than the state of the art, while outperforming all existing methods on standard benchmark datasets~\cite{Silberman:ECCV12,saxena2005learning}. We further propose a more efficient scheme for up-convolutions and combine it with the concept of residual learning~\cite{he2015deep} to create \emph{up-projection} blocks for the effective upsampling of feature maps. Last, we train the network by optimizing a loss based on the reverse Huber function (\emph{berHu})~\cite{zwald2012berhu} and demonstrate, both theoretically and experimentally, why it is beneficial and better suited for the task at hand.  
We thoroughly evaluate the influence of the network's depth, the loss function and the specific layers employed for up-sampling in order to analyze their benefits. Finally, to further assess the accuracy of our method, we employ the trained model within a 3D reconstruction scenario, in which we use a sequence of RGB frames and their predicted depth maps for Simultaneous Localization and Mapping (SLAM).

\section{Related Work}

\begin{figure*}[t]
	\centering
	\includegraphics[width=0.95\linewidth]{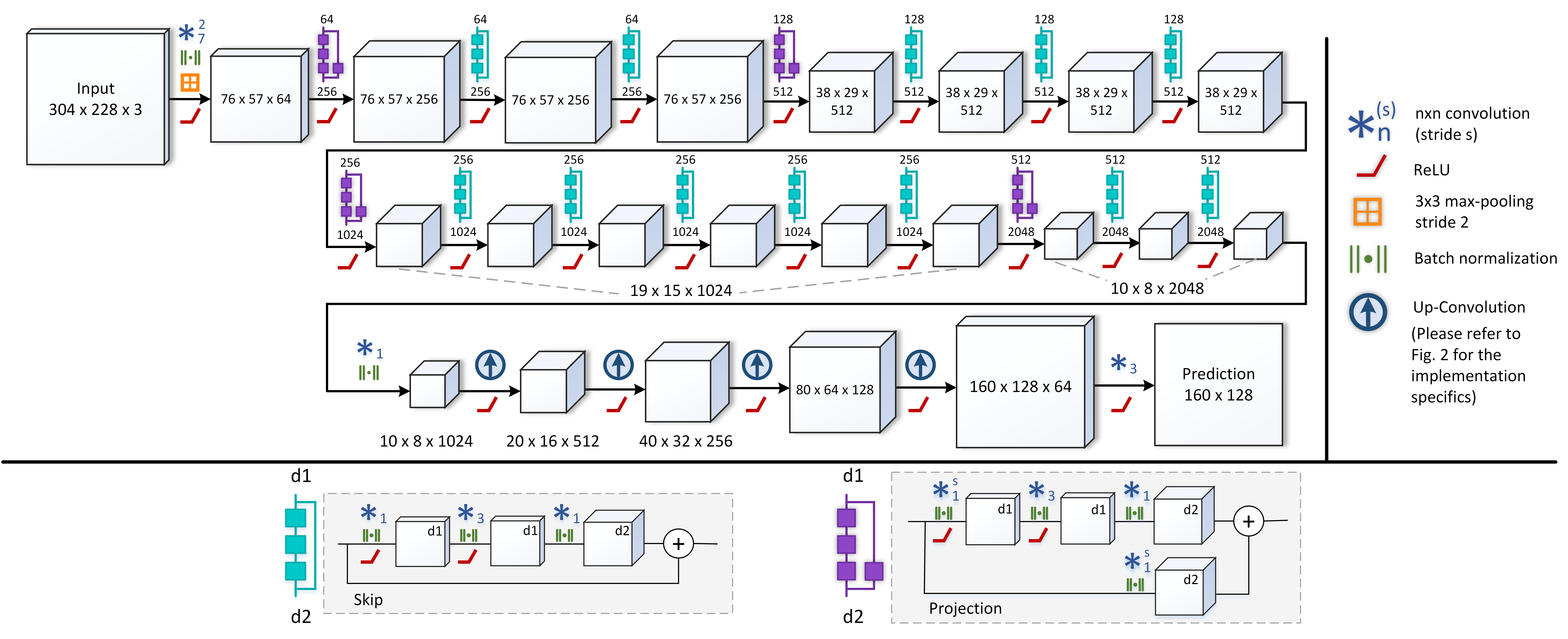}
	\caption{\textbf{Network architecture.} The proposed architecture builds upon ResNet-50. We replace the fully-connected layer, which was part of the original architecture, with our novel up-sampling blocks, yielding an output of roughly half the input resolution} 
	\label{fig:network}
	\vspace{-0.5em}
\end{figure*}

Depth estimation from image data has originally relied on stereo vision~\cite{memisevic2011stereopsis,sinz2004learning}, using image pairs of the same scene to reconstruct 3D shapes. In the single-view case, most approaches relied on motion (Structure-from-Motion~\cite{szeliski11}) or different shooting conditions (Shape-from-Shading~\cite{zhang1999shape}, Shape-from-Defocus~\cite{suwajanakorn2015}). Despite the ambiguities that arise in lack of such information, but inspired by the analogy to human depth perception from monocular cues, depth map prediction from a \emph{single} RGB image has also been investigated. Below, we focus on the related work for single RGB input, similar to our method.

Classic methods on monocular depth estimation have mainly relied on hand-crafted features and used probabilistic graphical models to tackle the problem~\cite{hoiem2005geometric,liu2010single,saxena2005learning,Saxena09}, usually making strong assumptions about scene geometry. One of the first works, by Saxena \etal~\cite{saxena2005learning}, uses a MRF to infer depth from local and global features extracted from the image, while superpixels~\cite{achanta2012slic} are introduced in the MRF formulation in order to enforce neighboring constraints. Their work has been later extended to 3D scene reconstruction~\cite{Saxena09}. Inspired by this work, Liu \etal~\cite{liu2010single} combine the task of semantic segmentation with depth estimation, where predicted labels are used as additional constraints to facilitate the optimization task. Ladicky \etal~\cite{Ladicky14} instead jointly predict labels and depths in a classification approach. 

A second cluster of related work comprises non-parametric approaches for depth transfer~\cite{Karsch12,Konrad12,liu2011sift,Liu14}, which typically perform feature-based matching (\eg GIST~\cite{Oliva01}, HOG~\cite{Dalal05}) between a given RGB image and the images of a RGB-D repository in order to find the nearest neighbors; the retrieved depth counterparts are then warped and combined to produce the final depth map. Karsch \etal~\cite{Karsch12} perform warping using SIFT Flow~\cite{liu2011sift}, followed by a global optimization scheme, whereas Konrad \etal~\cite{Konrad12} compute a median over the retrieved depth maps followed by cross-bilateral filtering for smoothing. Instead of warping the candidates, Liu \etal~\cite{Liu14}, formulate the optimization problem as a Conditional Random Field (CRF) with continuous and discrete variable potentials. Notably, these approaches rely on the assumption that similarities between regions in the RGB images imply also similar depth cues. 

More recently, remarkable advances in the field of deep learning drove research towards the use of CNNs for depth estimation. Since the task is closely related to semantic labeling, most works have built upon the most successful architectures of the ImageNet Large Scale Visual Recognition Challenge (ILSVRC)~\cite{ILSVRC15}, often initializing their networks with AlexNet~\cite{krizhevsky2012imagenet} or the deeper VGG~\cite{simonyan2014very}. Eigen \etal~\cite{Eigen14} have been the first to use CNNs for regressing dense depth maps from a single image in a two-scale architecture, where the first stage -- based on AlexNet -- produces a coarse output and the second stage refines the original prediction. Their work is later extended to additionally predict normals and labels with a deeper and more discriminative model -- based on VGG -- and a three-scale architecture for further refinement~\cite{Eigen15}. Unlike the deep architectures of~\cite{Eigen15,Eigen14}, Roy and Todorovic~\cite{roy16monocular} propose combining CNNs with regression forests, using very shallow architectures at each tree node, thus limiting the need for big data. 

Another direction for improving the quality of the predicted depth maps has been the combined use of CNNs and graphical models~\cite{Li15,Liu15,wang2015towards}. Liu \etal~\cite{Liu15} propose to learn the unary and pairwise potentials during CNN training in the form of a CRF loss, while Li \etal~\cite{Li15} and Wang \etal~\cite{wang2015towards} use hierarchical CRFs to refine their patch-wise CNN predictions from superpixel down to pixel level.

Our method uses a CNN for depth estimation and differs from previous work in that it improves over the typical fully-connected layers, which are expensive with respect to the number of parameters, with a fully convolutional model incorporating efficient residual up-sampling blocks, that we refer to as \emph{up-projections} and which prove to be more suitable when tackling high-dimensional regression problems. 

\section{Methodology}


In this section, we describe our model for depth prediction from a single RGB image. We first present the employed architecture, then analyze the new components proposed in this work. Subsequently, we propose a loss function suitable for the optimization of the given task.

\subsection{CNN Architecture}
\label{sec:architecture}
Almost all current CNN architectures contain a contractive part that progressively decreases the input image resolution through a series of convolutions and pooling operations, giving higher-level neurons large receptive fields, 
thus capturing more global information. 
In regression problems in which the desired output is a high resolution image, some form of up-sampling is required in order to obtain a larger output map. Eigen \etal~\cite{Eigen15,Eigen14}, use fully-connected layers as in a typical classification network, 
yielding a full receptive field. The outcome is then reshaped to the output resolution. 

We introduce a fully convolutional network for depth prediction. Here, the receptive field is an important aspect of the architectural design, as there are no explicit full connections. 
Specifically, assume we set an input of $304 \times 228$ pixels (as in~\cite{Eigen14}) and predict an output map that will be at approximately half the input resolution. We investigate popular architectures (AlexNet~\cite{krizhevsky2012imagenet}, VGG-16~\cite{simonyan2014very}) as the contractive part, since their pre-trained weights facilitate convergence. 
The receptive field at the last convolutional layer of AlexNet is $151\times151$ pixels, allowing only very low resolution input images when true global information (\eg monocular cues) should be captured by the network without fully-connected layers.
A larger receptive field of $276\times276$ is achieved by VGG-16 but still sets a limit to the input resolution. 
Eigen and Fergus~\cite{Eigen15} show a substantial improvement when switching from AlexNet to VGG, but since both their models use fully-connected layers, this is due to the higher discriminative power of VGG.

Recently, ResNet \cite{he2015deep} introduced skip layers that by-pass two or more convolutions and are summed to their output, including batch normalization~\cite{ioffe2015batch} after every convolution (see Fig.~\ref{fig:network}). Following this design, it is possible to create much deeper networks without facing degradation or vanishing gradients. Another beneficial property of these extremely deep architectures is their large receptive field; ResNet-50 captures input sizes of $483\times483$, large enough to fully capture the input image  even in higher resolutions.
Given our input size and this architecture, the last convolutional layers result in 2048 feature maps of spatial resolution $10 \times 8$ pixels, when removing the last pooling layer. As we show later, the proposed model, which uses residual up-convolutions, produces an output of $160 \times 128$ pixels. If we instead added a fully-connected layer of the same size, it would introduce $3.3$ billion parameters, worth $12.6$GB in memory, rendering this approach impossible on current hardware. This further motivates our proposal of a fully convolutional architecture with up-sampling blocks that contain fewer weights while improving the accuracy of the predicted depth maps. 

Our proposed architecture can be seen in Fig.~\ref{fig:network}. The feature map sizes correspond to the network trained for input size $304\times228$, in the case of NYU Depth v2 data set~\cite{Silberman:ECCV12}. The first part of the network is based on ResNet-50 and initialized with pre-trained weights.
The second part of our architecture guides the network into learning its upscaling through a sequence of unpooling and convolutional layers. Following the set of these up-sampling blocks, dropout is applied and succeeded by a final convolutional layer yielding the  prediction. 
\vspace{-0.5em}

\paragraph{Up-Projection Blocks.}
 
\begin{figure}[t]
	\centering
	\includegraphics[width=0.95\linewidth]{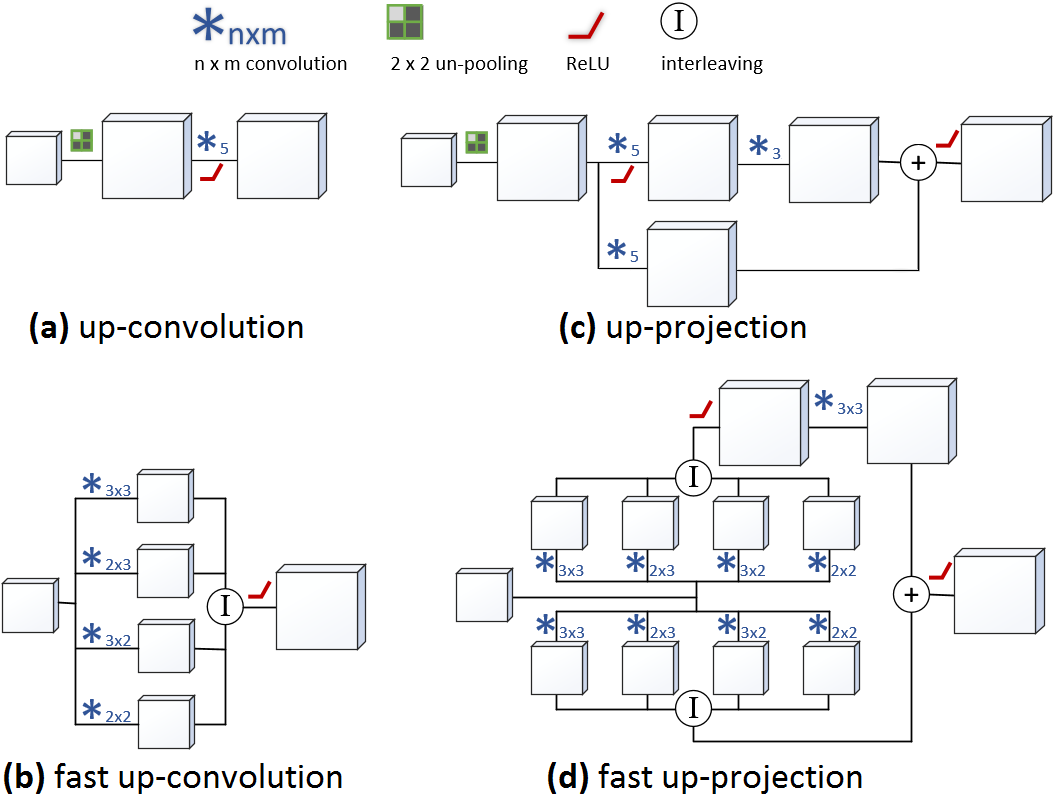}
	\caption{\textbf{From up-convolutions to up-projections.} \textbf{(a)} Standard up-convolution. \textbf{(b)} The equivalent but faster up-convolution. \textbf{(c)} Our novel up-projection block, following residual logic. \textbf{(d)} The faster equivalent version of (c)} 
	\label{fig:upprojection}
	\vspace{-1em}
\end{figure}

Unpooling layers~\cite{dosovitskiy2015learning,long2015fully,zeiler2011adaptive}, perform the reverse operation of pooling, increasing the spatial resolution of feature maps. We adapt the approach described in~\cite{dosovitskiy2015learning} for the implementation of unpooling layers, in order to double the size by mapping each entry into the top-left corner of a $2\times2$ (zero) kernel. Each such layer is followed by a $5\times5$ convolution -- so that it is applied to more than one non-zero elements at each location -- and successively by ReLU activation. We refer to this block as up-convolution. 
Empirically, we stack four such up-convolutional blocks, \ie 16x upscaling of the smallest feature map, resulting in the best trade-off between memory consumption and resolution. We found that performance did not increase when adding a fifth block. 

We further extend simple up-convolutions using a similar but inverse concept to \cite{he2015deep} to create up-sampling res-blocks. The idea is to introduce a simple $3\times3$ convolution after the up-convolution and to add a \emph{projection} connection from the lower resolution feature map to the result, as shown in Fig.~\ref{fig:upprojection}(c). Because of the different sizes, the small-sized map needs to be up-sampled using another up-convolution in the projection branch, but since the unpooling only needs to be applied once for both branches, we just apply the $5\times5$ convolutions separately on the two branches. We call this new up-sampling block \emph{up-projection} since it extends the idea of the projection connection~\cite{he2015deep} to up-convolutions. Chaining up-projection blocks allows high-level information to be more efficiently passed forward in the network while progressively increasing feature map sizes. This enables the construction of our coherent, fully convolutional network for depth prediction. 
Fig.~\ref{fig:upprojection} shows the differences between an up-convolutional block to up-projection block. It also shows the corresponding fast versions that will be described in the following section. 
\vspace{-0.5em}

\paragraph{Fast Up-Convolutions.}

\begin{figure}[ht]
	\centering
	\includegraphics[width=0.99\linewidth]{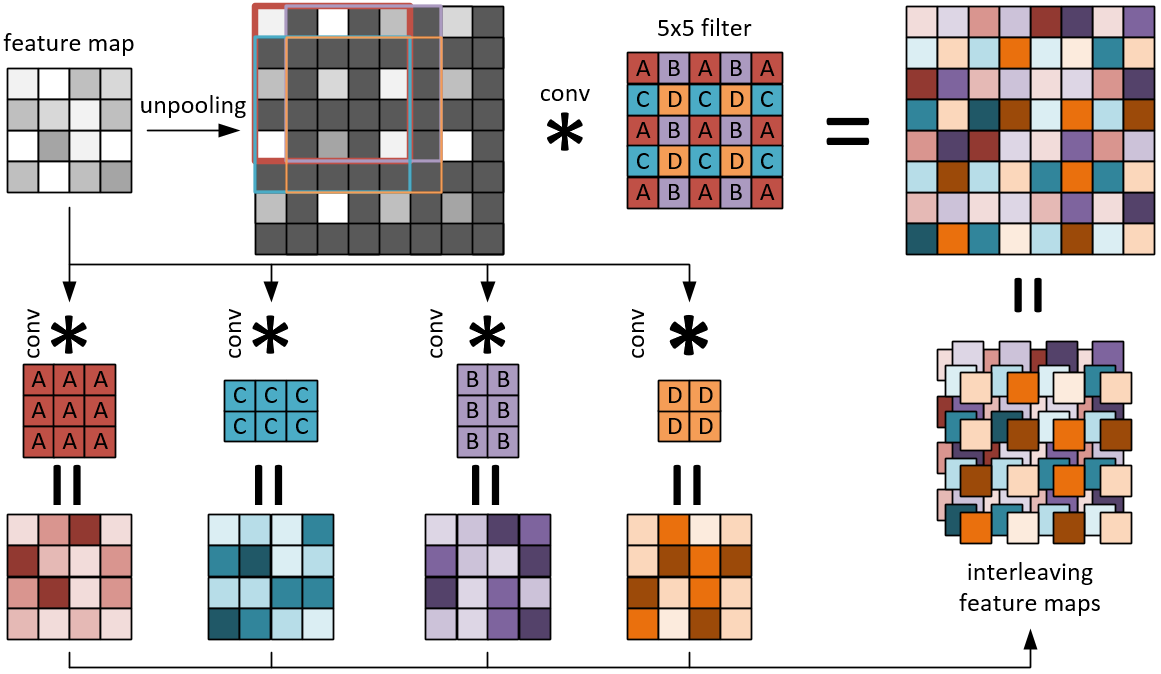}
	\caption{\textbf{Faster up-convolutions.} Top row: the common up-convolutional steps: unpooling doubles a feature map's size, filling the holes with zeros, and a $5\times5$ convolution filters this map. Depending on the position of the filter, only certain parts of it (A,B,C,D) are multiplied with non-zero values. This motivates convolving the original feature map with the 4 differently composed filters (bottom part) and interleaving them to obtain the same output, while avoiding zero multiplications. A,B,C,D only mark locations and the actual weight values will differ} 
	\label{fig:unpooling}
	\vspace{-0.5em}
\end{figure}

One further contribution of this work is to reformulate the up-convolution operation so to make it more efficient, leading to a decrease of training time of the whole network of around 15\%. This also applies to the newly introduced up-projection operation.
The main intuition is as follows: after unpooling 75\% of the resulting feature maps contain zeros, thus the following $5\times5$ convolution mostly operates on zeros which can be avoided in our modified formulation. 
This can be observed in Fig.~\ref{fig:unpooling}. In the top left the original feature map is unpooled (top middle) and then convolved by a $5\times5$ filter. We observe that in an unpooled feature map, depending on the location (red, blue, purple, orange bounding boxes) of the $5\times5$ filter, only certain weights are multiplied with potentially non-zero values. These weights fall into four non-overlapping groups, indicated by different colors and A,B,C,D in the figure. Based on the filter groups, we arrange the original $5\times5$ filter to four new filters of sizes (A) $3\times3$, (B) $3\times2$, (C) $2\times3$ and (D) $2\times2$. Exactly the same output as the original operation (unpooling and convolution) can now be achieved by interleaving the elements of the four resulting feature maps as in Fig.~\ref{fig:unpooling}.
The corresponding changes from a simple up-convolutional block to the proposed up-projection are shown in Fig.~\ref{fig:upprojection} (d). 

\subsection{Loss Function} 
\label{scc:loss}


A standard loss function for optimization in regression problems is the $\mathcal{L}_2$ loss, minimizing the squared euclidean norm between predictions $\tilde{y}$ and ground truth $y$: $\mathcal{L}_2(\tilde{y} - y) = ||\tilde{y} - y||_2^2$. Although this produces good results in our test cases, we found that using the reverse Huber (berHu)~\cite{owen2007robust,zwald2012berhu} as loss function $\mathcal{B}$ yields a better final error than $\mathcal{L}_2$.
\begin{equation}
\mathcal{B}(x) = \begin{cases} 
|x| & |x| \leq c, \\
\frac{x^2 + c^2}{2c} & |x| > c. \\
\end{cases}
\end{equation}
The Berhu loss is equal to the $\mathcal{L}_1(x) = |x|$ norm when $x \in [-c, c]$ and equal to $\mathcal{L}_2$ outside this range. The version used here is continuous and first order differentiable at the point $c$ where the switch from $\mathcal{L}_1$ to $\mathcal{L}_2$ occurs. 
In every gradient descent step, when we compute $\mathcal{B}(\tilde{y} - y)$ we set $c = \frac{1}{5} \max_i(|\tilde{y}_i - y_i|)$, where $i$ indexes all pixels over each image in the current batch, that is $20\%$ of the maximal per-batch error. Empirically, BerHu shows a good balance between the two norms in the given problem; it puts high weight towards samples/pixels with a high residual because of the $\mathcal{L}_2$ term, contrary for example to a robust loss, such as Tukey's biweight function that ignores samples with high residuals~\cite{belagiannis2015robust}. At the same time, $\mathcal{L}_1$ accounts for a greater impact of smaller residuals' gradients than $\mathcal{L}_2$ would. 

We provide two further intuitions with respect to the difference between $\mathcal{L}_2$ and berHu loss. In both datasets that we experimented with, we observe a heavy-tailed distribution of depth values, also reported in~\cite{roy16monocular}, for which Zwald and Lambert-Lacroix~\cite{zwald2012berhu} show that the berHu loss function is more appropriate. This could also explain why \cite{Eigen15,Eigen14} experience better convergence when predicting the log of the depth values, effectively moving a log-normal distribution back to Gaussian. Secondly we see the greater benefit of berHu in the small residuals during training as there the $\mathcal{L}_1$ derivative is greater than $\mathcal{L}_2$'s. This manifests in the error measures rel.~and $\delta_1$ (Sec.~\ref{sec:results}), which are more sensitive to small errors. 
\section{Experimental Results}
\label{sec:results}

In this section, we provide a thorough analysis of our methods, evaluating the different components that comprise the down-sampling and up-sampling part of the CNN architecture. We also report the quantitative and qualitative results obtained by our model and compare to the state of the art in two standard benchmark datasets for depth prediction, \ie NYU Depth v2~\cite{Silberman:ECCV12} (indoor scenes) and Make3D~\cite{Saxena09} (outdoor scenes). 
\begin{table}
	\centering
	\def\arraystretch{1.3}
	\resizebox{\linewidth}{!}{
		\begin{tabular}{ | l l c | r || r r r r r r |}
			\hline
			Architecture & \hspace{1.7cm} & Loss & \multicolumn{1}{c||}{\#params} & \multicolumn{1}{c}{rel} & \multicolumn{1}{c}{rms} & \multicolumn{1}{c}{$\log_{10}$} & \multicolumn{1}{c}{$\delta_1$} & \multicolumn{1}{c}{$\delta_2$} & \multicolumn{1}{c|}{$\delta_3$} \\ \hline \hline
			AlexNet & FC & $\mathcal{L}_2$ & $104.4\times10^6$ & 0.209 & 0.845 & 0.090 & 0.586 & 0.869 & 0.967 \\
			&  & berHu &  & 0.207 & 0.842 & 0.091 & 0.581 & 0.872 & 0.969 \\
			& UpConv & $\mathcal{L}_2$ & $6.3\times10^6$ & 0.218 & 0.853 & 0.094 & 0.576 & 0.855 & 0.957 \\ 
			&  & berHu &  & 0.215 & 0.855 & 0.094 & 0.574 & 0.855 & 0.958 \\ \hline 
			VGG	& UpConv & $\mathcal{L}_2$ & $18.5\times10^6$ & 0.194 & 0.746 & 0.083 & 0.626 & 0.894 & 0.974 \\ 
			&  & berHu & & 0.194 & 0.790 & 0.083 & 0.629 & 0.889 & 0.971 \\ \hline 
			ResNet & FC-160x128 & berHu & $359.1\times10^6$ & 0.181 & 0.784 & 0.080 & 0.649 & 0.894 & 0.971 \\ 
			& FC-64x48 & berHu & $73.9\times10^6$ & 0.154 & 0.679 & 0.066 & 0.754 & 0.938 & 0.984 \\ 
			& DeConv & $\mathcal{L}_2$ & $28.5\times10^6$& 0.152 & 0.621 & 0.065 & 0.749 & 0.934 & 0.985 \\ 
			& UpConv & $\mathcal{L}_2$ & $43.1\times10^6$ & 0.139 & 0.606 & 0.061 & 0.778 & 0.944 & 0.985 \\ 
			&  & berHu & & 0.132 & 0.604 & 0.058 & 0.789 & 0.946 & 0.986 \\ 
			& UpProj & $\mathcal{L}_2$ & $63.6\times10^6$ & 0.138 & 0.592 & 0.060 & 0.785 & 0.952 & 0.987 \\ 
			& & berHu & & \textbf{0.127} & \textbf{0.573} & \textbf{0.055} & \textbf{0.811} & \textbf{0.953} & \textbf{0.988} \\ \hline
			\noalign{\smallskip}
		\end{tabular}}
		\caption{Comparison of the proposed approach against different variants on the NYU Depth v2 dataset. For the reported errors rel, rms, $\log_{10}$ lower is better, whereas for the accuracies $\delta_i < 1.25^i$ higher is better}
		\label{tab:resVariantsNYU}
		\vspace{-1em}
	\end{table}

\subsection{Experimental Setup}
For the implementation of our network we use \mbox{MatConvNet}~\cite{vedaldi15matconvnet}, and train on a single NVIDIA GeForce GTX TITAN with 12GB of GPU memory. Weight layers of the down-sampling part of the architecture are initialized by the corresponding models (AlexNet, VGG, ResNet) pre-trained on the ILSVRC~\cite{ILSVRC15} data for image classification. Newly added layers of the up-sampling part are initialized as random filters sampled from a normal distribution with zero mean and 0.01 variance.

The network is trained on RGB inputs to predict the corresponding depth maps. We use data augmentation to increase the number of training samples. The input images and corresponding ground truth are transformed using small rotations, scaling, color transformations and flips with a 0.5 chance, with values following Eigen \etal~\cite{Eigen14}. Finally, we model small translations by random crops of the augmented images down to the chosen input size of the network. 

For the quantitative evaluation that follows, the same error metrics which have been used in prior works~\cite{Eigen15,Eigen14,Ladicky14,Li15,Liu15} are computed on our experimental results. 

%
%
%
%

\subsection{NYU Depth Dataset}

\begin{table}[t]
	\def\arraystretch{1.3}
	\centering
	\resizebox{\linewidth}{!}{
	\begin{tabular}{ | l || *{7}{c |} }
		\hline
		\textbf{NYU Depth v2} & rel & rms & rms(log) & $\log_{10}$ & $\delta_1$ & $\delta_2$ & $\delta_3$ \\ \hline \hline
		Karsch \etal~\cite{Karsch12} & 0.374 & 1.12 & - & 0.134 & - & - & - \\ \hline
		Ladicky \etal~\cite{Ladicky14} & - & - & - & -  & 0.542 & 0.829 & 0.941 \\ \hline
		Liu \etal~\cite{Liu14} & 0.335 & 1.06 & - & 0.127 & - & - & - \\ \hline
		Li \etal~\cite{Li15} & 0.232 & 0.821 & - & 0.094 & 0.621 & 0.886 & 0.968 \\ \hline
		Liu \etal~\cite{Liu15} & 0.230 & 0.824 & - & 0.095 & 0.614 & 0.883 & 0.971 \\ \hline
		Wang \etal~\cite{wang2015towards} & 0.220 & 0.745 & 0.262 & 0.094 & 0.605 & 0.890 & 0.970 \\ \hline
		Eigen \etal~\cite{Eigen14} & 0.215 & 0.907  & 0.285 & - & 0.611 & 0.887 & 0.971 \\ \hline
		Roy and Todorovic~\cite{roy16monocular} & 0.187 & 0.744 & - & 0.078 & - & - & - \\ \hline
		Eigen and Fergus~\cite{Eigen15} & 0.158 & 0.641 & 0.214 & - & 0.769 & 0.950 & \textbf{0.988} \\ \hline \hline
		ours (ResNet-UpProj) & \textbf{0.127} & \textbf{0.573} & \textbf{0.195} & \textbf{0.055} & \textbf{0.811} & \textbf{0.953} & \textbf{0.988} \\ \hline 
		\noalign{\smallskip}
	\end{tabular}}
	\caption{Comparison of the proposed approach against the state of the art on the NYU Depth v2 dataset. The values are those originally reported by the authors in their respective paper
	\label{tab:resNYU}}
	\vspace{-0.8em}
\end{table}

\begin{figure*}[ht]
	\centering
	\includegraphics[width=0.95\linewidth]{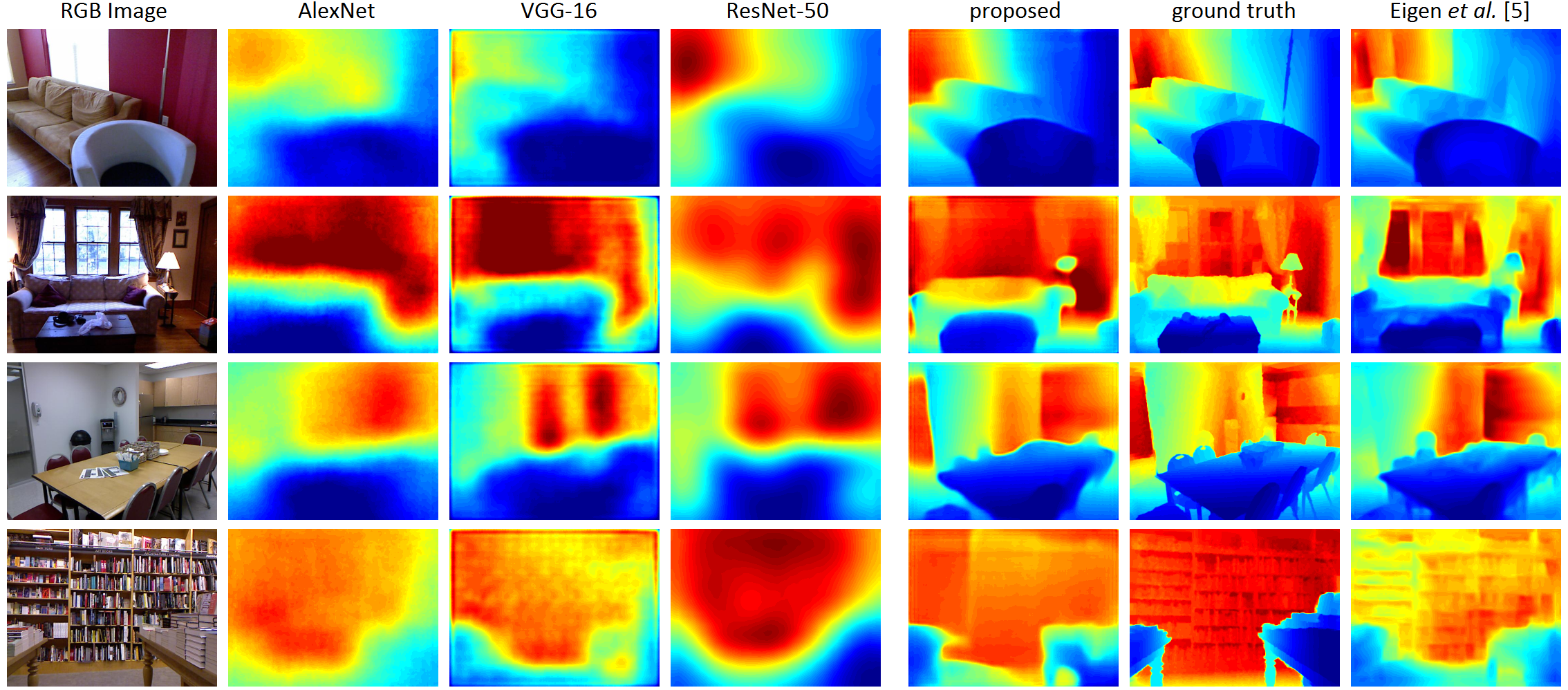}
	\caption{\textbf{Depth Prediction on NYU Depth} Qualitative results showing predictions using AlexNet, VGG, and the fully-connected ResNet compared to our model and the predictions of~\cite{Eigen15}. All colormaps are scaled equally for better comparison} 
	\label{fig:comparison}
	\vspace{-0.7em}
\end{figure*}

First, we evaluate on one of the largest RGB-D data sets for indoor scene reconstruction, NYU Depth v2~\cite{Silberman:ECCV12}. The raw dataset consists of 464 scenes, captured with a Microsoft Kinect, with the official split consisting in 249 training and 215 test scenes. For training, however, our method only requires a small subset of the raw distribution. We sample equally-spaced frames out of each training sequence, resulting in approximately 12k unique images. After offline augmentations of the extracted frames, our dataset comprises approximately 95k pairs of RGB-D images. We point out that our dataset is radically smaller than that required to train the model in~\cite{Eigen15,Eigen14}, consisting of 120k unique images, as well as the 800k samples extracted in the patch-wise approach of~\cite{Li15}. Following \cite{Eigen14}, the original frames of size $640 \times 480$ pixels are down-sampled to $1/2$ resolution and center-cropped to $304 \times 228$ pixels, as input to the network. 
At last, we train our model with a batch size of 16 for approximately 20 epochs. The starting learning rate is $10^{-2}$ for all layers, which we gradually reduce every 6-8 epochs, when we observe plateaus; momentum is 0.9. 

For the quantitative evaluation of our methods and comparison to the state of the art on this data set, we compute various error measures on the commonly used test subset of 654 images. 
The predictions' size depends on the specific model; in our configuration, which consists of four up-sampling stages, the corresponding output resolutions are $128 \times 96$ for AlexNet, $144 \times 112$ for VGG and $160 \times 128$ for ResNet-based models. The predictions are then up-sampled back to the original size ($640 \times 480$) using bilinear interpolation and compared against the provided ground truth with filled-in depth values for invalid pixels. 
\vspace{-1em}

\paragraph{Architecture Evaluation.}

In Table~\ref{tab:resVariantsNYU} we compare different CNN variants of the proposed architecture, in order to study the effect of each component.  
First, we evaluate the influence of the depth of the architecture using the convolutional blocks of AlexNet, VGG-16 and ResNet-50. It becomes apparent that a fully convolutional architecture (UpConv) on AlexNet is outperformed by the typical network with full connections (FC). As detailed in Sec.~\ref{sec:architecture}, a reason for this is the relatively small field of view in AlexNet, which is not enough to capture global information that is needed when removing the fully-connected layers. Instead, using VGG as the core architecture, improves the accuracy on depth estimation. As a fully-connected VGG variant for high-dimensional regression would incorporate a high number of parameters, we only perform tests on the fully convolutional (UpConv) model here. However, a VGG-based model with fully-connected layers was indeed employed by~\cite{Eigen15} (for their results see Table~\ref{tab:resNYU}) performing better than our fully convolutional VGG-variant mainly due to their multi-scale architecture, including the refinement scales. 

Finally, switching to ResNet with a fully-connected layer (ResNet-FC) -- without removing the final pooling layer -- achieves similar performance to~\cite{Eigen15} for a low resolution output ($64\times48$), using 10 times fewer data; however increasing the output resolution ($160\times128$) results in such a vast number of parameters that convergence becomes harder. 
This further motivates the reasoning for the replacement of fully-connected layers and the need for more efficient upsampling techniques, when dealing with high-dimensional problems.
Our fully convolutional variant using simple up-convolutions (ResNet-UpConv) improves accuracy, and at last, the proposed architecture (ResNet-UpProj), enhanced with the up-projection blocks, gives by far the best results. As far as the number of parameters is concerned, we see a drastic decrease when switching from fully-connected layers to fully convolutional networks. Another common up-sampling technique that we investigated is deconvolution with successive $2\times2$ kernels, but the up-projections notably outperformed it. Qualitatively, since our method consists in four successive up-sampling steps (2x resolution per block), it can preserve more structure in the output when comparing to the FC-variant (see Fig.~\ref{fig:comparison}).

In all shown experiments the berHu loss outperforms $\mathcal{L}_2$. The difference is higher in relative error which can be explained by the larger gradients of $\mathcal{L}_1$ (berHu) over $\mathcal{L}_2$ for small residuals; the influence on the relative error is higher, as there pixels in smaller distances are more sensitive to smaller errors. This effect is also well visible as a stronger gain in the challenging $\delta_1$ measure.

Finally, we measure the timing of a single up-convolutional block for a single image (1.5 ms) and compare to our up-projection (0.14 ms). This exceeds the theoretical speed up of 4 and is due to the fact that smaller filter sizes benefit more from the linearization inside cuDNN. Furthermore, one of the advantages of our model is the overall computation time. Predicting the depth map of a single image takes only 55ms with the proposed up-sampling (78ms with up-convolutions) on our setup. This enables real-time processing images, for example from a web-cam. Further speed up can be achieved when several images are processed in a batch. A batch size of 16 results in 14ms per image with up-projection and 28ms for up-convolutions.
\vspace{-1em} 

\begin{figure}[t]
	\centering
	\includegraphics[width=0.8\linewidth]{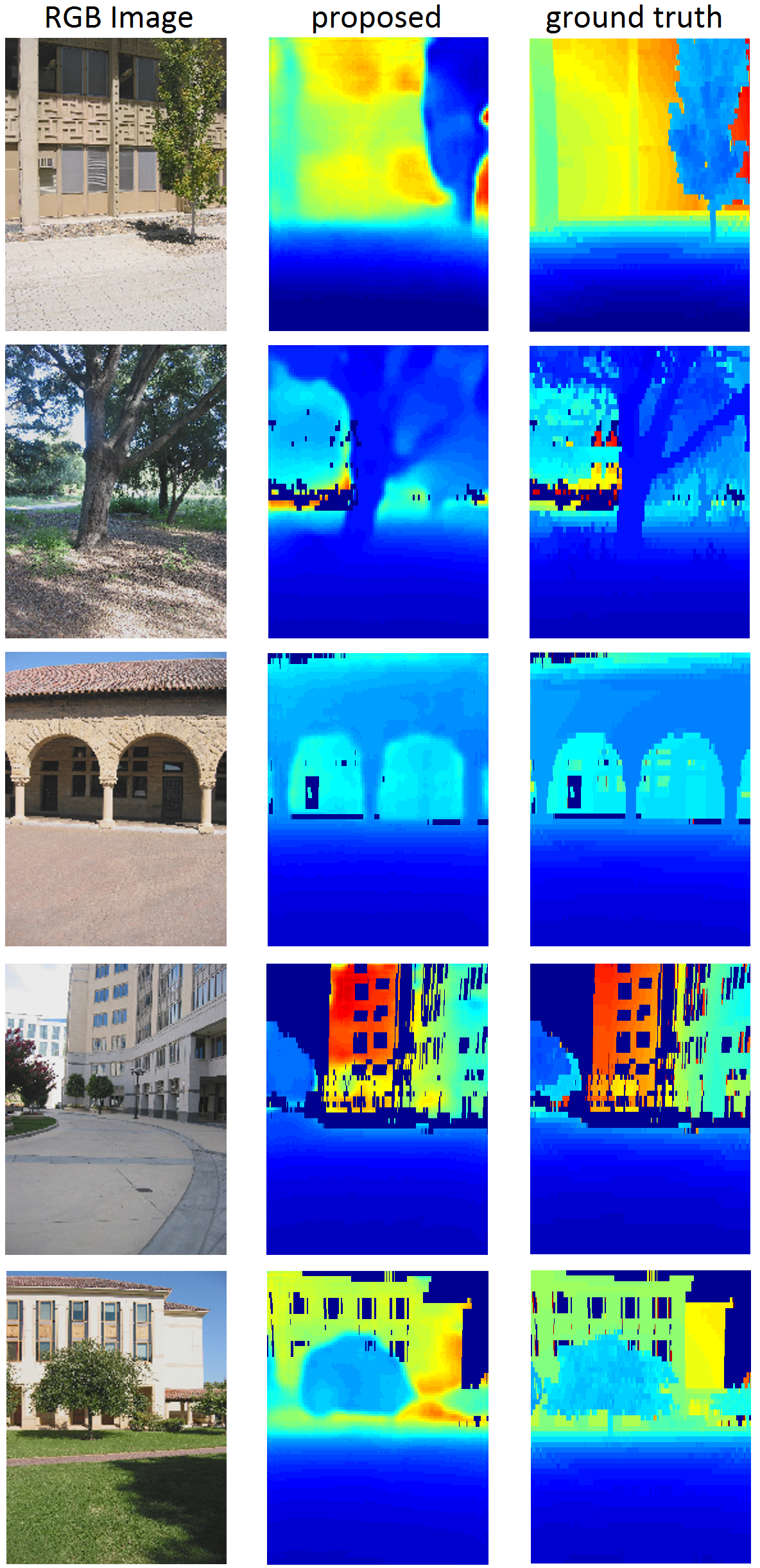}
	\caption{\textbf{Depth Prediction on Make3D.} Displayed are RGB images (first row), ground truth depth maps (middle row) and our predictions (last row). Pixels that correspond to distances $>70m$ in the ground truth are masked out}	
	\label{fig:make3d}
	\vspace{-1.1em}
\end{figure}

\paragraph{Comparison with related methods.}

\begin{figure*}[t]
	\centering
	\includegraphics[width=0.95\linewidth]{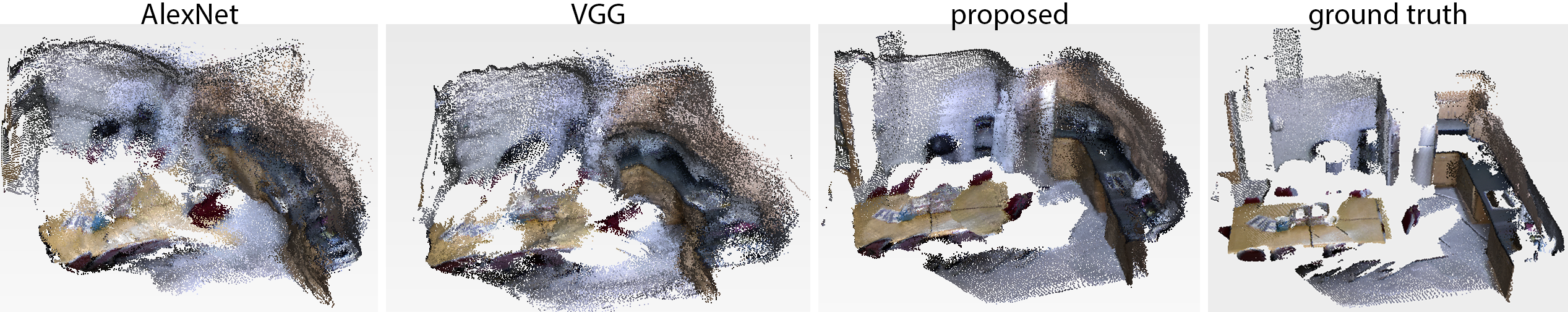}
	\caption{\textbf{3D SLAM} Comparison of the 3D reconstructions obtained on NYU Depth dataset between the ground-truth depth (left-most) and the depth predicted, respectively (left to right), by AlexNet, VGG and our architecture.} 
	\label{fig:slam}
	\vspace{-0.5em}
\end{figure*}

In Table~\ref{tab:resNYU} we compare the results obtained by the proposed architecture to those reported by related work. 
Additionally, in Fig.~\ref{fig:comparison} we qualitatively compare the accuracy of the estimated depth maps using the proposed approach (ResNet-UpProj) with that of the different variants (AlexNet, VGG, ResNet-FC-64x48) as well as with the publicly available predictions of Eigen and Fergus~\cite{Eigen15}. One can clearly see the improvement in quality from AlexNet to ResNet, however the fully-connected variant of ResNet, despite its increased accuracy, is still limited to coarse predictions. The proposed fully convolutional model greatly improves edge quality and structure definition in the predicted depth maps.

Interestingly, our depth predictions exhibit noteworthy visual quality, even though they are derived by a single model, trained end-to-end, without any additional post-processing steps, as for example the CRF inference of~\cite{Li15,wang2015towards}. On the other hand, \cite{Eigen15} refine their predictions through a multi-scale architecture that combines the RGB image and the original prediction to create visually appealing results. However, they sometimes mis-estimate the global scale (second and third row) or introduce noise in case of highly-textured regions in the original image, even though there is no actual depth border in the ground truth (last row). Furthermore, we compare to the number of parameters in \cite{Eigen15}, which we calculated as $218$ million for the three scales, that is approximately $3.5$ times more than our model. Instead, the CNN architecture proposed here is designed with feasibility in mind; the number of parameters should not increase uncontrollably in high-dimensional problems. 
This further means a reduction in the number of gradient steps required as well as the data samples needed for training. Our single network generalizes better and successfully tackles the problem of coarseness that has been encountered by previous CNN approaches on depth estimation. 

\subsection{Make3D Dataset}

\begin{table}
	\centering
	\def\arraystretch{1.05}
	\begin{center}
		\begin{tabular}{ | l || c | c | c | }
			\hline
			\textbf{Make3D} & rel & rms & $\log_{10}$ \\ \hline \hline
			Karsch \etal~\cite{Karsch12} & 0.355 & 9.20 & 0.127 \\ \hline
			Liu \etal~\cite{Liu14} & 0.335 & 9.49 & 0.137  \\ \hline
			Liu \etal~\cite{Liu15} & 0.314 & 8.60 & 0.119\\ \hline
			Li \etal~\cite{Li15} & 0.278 & 7.19 & 0.092 \\ \hline \hline
			ours ($\mathcal{L}_2$) & 0.223 & 4.89 & 0.089 \\ \hline
			ours (berHu) & \textbf{0.176} & \textbf{4.46} & \textbf{0.072}  \\ 
			\hline 
		\end{tabular}
	\end{center}
	\vspace{-0.5em}
	\caption{\textbf{Comparison with the state of the art.} We report our results with l2 and berHu loss. The shown values of the evaluated methods are those reported by the authors in their paper	\label{tab:resMAKE3D}}
	\vspace{-1em}
\end{table}

In addition, we evaluated our model on Make3D data set~\cite{Saxena09} of outdoor scenes. It consists of 400 training and 134 testing images, gathered using a custom 3D scanner. As the dataset acquisition dates to several years ago, the ground truth depth map resolution is restricted to $305\times55$, unlike the original RGB images of $1704 \times 2272$ pixels. Following~\cite{Liu14}, we resize all images to $345\times460$ and further reduce the resolution of the RGB inputs to the network by half because of the large architecture and hardware limitations. We train on an augmented data set of around 15k samples using the best performing model (ResNet-UpProj) with a batch size of 16 images for 30 epochs. Starting learning rate is 0.01 when using the berHu loss, but it needs more careful adjustment starting at 0.005 when optimizing with $\mathcal{L}_2$. Momentum is 0.9. Please note that due to the limitations that come with the dataset, considering the low resolution ground truth and long range inaccuracies (\eg sky pixels mapped at 80m), we train against ground truth depth maps by masking out pixels of distances over 70m. 

In order to compare our results to state-of-the-art, we up-sample the predicted depth maps back to $345\times460$ using bilinear interpolation. Table~\ref{tab:resMAKE3D} reports the errors compared to previous work based on (C1) criterion, computed in regions of depth less than 70m as suggested by~\cite{Liu14} and as implied by our training. 
As an aside, \cite{Liu14} pre-process the images with a per-pixel sky classification to also exclude them from training. 
Our method significantly outperforms all previous works when trained with either $\mathcal{L}_2$ or berHu loss functions. In this challenging dataset, the advantage of berHu loss is more eminent. Also similarly to NYU, berHu improves the relative error more than the rms because of the weighting of close depth values. Qualitative results from this dataset are shown in Fig.~\ref{fig:make3d}. 

\subsection{Application to SLAM}
To complement the previous results, we demonstrate the usefulness of depth prediction within a SLAM application, with the goal of reconstructing the geometry of a 3D environment. In particular, we deploy a SLAM framework where frame-to-frame tracking is obtained via Gauss-Newton optimization on the pixelwise intensity differences computed on consecutive frame pairs as proposed in~\cite{Kerl13}, while fusion of depth measurements between the current frame and the global model is carried out via point-based fusion~\cite{Keller13}. 
We wish to point out that, to the best of our knowledge, this is the first demonstration of a SLAM reconstruction based on depth predictions from single images. 

A qualitative comparison between the SLAM reconstructions obtained using the depth values estimated with the proposed ResNet-UpProj architecture against that obtained using the ground truth depth values on part of a sequence of the NYU Depth dataset is shown in Fig.~\ref{fig:slam}. The figure also includes a comparison with the depth predictions obtained using AlexNet and VGG architectures. As it can be seen, the improved accuracy of the depth predictions, together with the good edge-preserving qualities of our up-sampling method, is not only noticeable in the qualitative results of Fig.~\ref{fig:comparison}, but also yields a much more accurate SLAM reconstruction compared to the other architectures. 
We wish to point out that, although we do not believe its accuracy could be yet compared to that achieved by methods exploiting temporal consistency for depth estimation such as SfM and monocular SLAM, our method does not explicitly rely on visual features to estimate depths, and thus holds the potential to be applied also on scenes characterized by low-textured surfaces such as walls, floors and other structures typically present in indoor environments. Although clearly outside the scope of this paper, we find these aspects relevant enough to merit future analysis.

\section{Conclusion}
In this work we present a novel approach to the problem of depth estimation from a single image. Unlike typical CNN approaches that require a multi-step process in order to refine their originally coarse depth predictions, our method  consists in a powerful, single-scale CNN architecture that follows residual learning. The proposed network is fully convolutional, comprising up-projection layers that allow for training much deeper configurations, while greatly reducing the number of parameters to be learned and the number of training samples required.
Moreover, we illustrate a faster and more efficient approach to up-convolutional layers. A thorough evaluation of the different architectural components has been carried out not only by optimizing with the typical l2 loss, but also with the berHu loss function, showing that it is better suited for the underlying value distributions of the ground truth depth maps. All in all, the model emerging from our contributions is not only simpler than existing methods, can be trained with less data in less time, but also achieves higher quality results that lead our method to state-of-the-art in two benchmark datasets for depth estimation. 



{\small
\bibliographystyle{ieee}
\bibliography{bibliography}
}

\end{document}